\documentclass[letterpaper, 10 pt, conference]{ieeeconf}  
\usepackage{geometry}
\IEEEoverridecommandlockouts                              

\usepackage{overpic} 
\graphicspath{{bilder/}}
\usepackage{xcolor}
\usepackage{float}
\usepackage{multirow}
\usepackage{amssymb}
\usepackage{booktabs}
\usepackage{amsmath}

\title{\LARGE \bf 
DVMN: Dense Validity Mask Network for Depth Completion
}

\author{Laurenz Reichardt$^{1}$, Patrick Mangat$^{1}$ and Oliver Wasenm\"uller$^{1}$
\thanks{$^{1}$Mannheim University of Applied Sciences, Germany.\linebreak{\tt\small l.reichardt@stud.hs-mannheim.de}\linebreak{\tt\small p.mangat@hs-mannheim.de}\linebreak{\tt\small o.wasenmueller@hs-mannheim.de}}}

\begin{document}

\maketitle
\thispagestyle{empty}
\pagestyle{empty}


\begin{abstract}
LiDAR depth maps provide environmental guidance in a variety of applications. However, such depth maps are typically sparse and insufficient for complex tasks such as autonomous navigation. State of the art methods use image guided neural networks for dense depth completion. We develop a guided convolutional neural network focusing on gathering dense and valid information from sparse depth maps. To this end, we introduce a novel layer with spatially variant and content-depended dilation to include additional data from sparse input. Furthermore, we propose a sparsity invariant residual bottleneck block. We evaluate our Dense Validity Mask Network (DVMN) on the KITTI depth completion benchmark and achieve state of the art results. At the time of submission, our network is the leading method using sparsity invariant convolution.
\end{abstract}

\section{INTRODUCTION}
\begin{figure}[t!]
	\centering
	\begin{overpic}[width=0.85\columnwidth, ,tics=5]
		{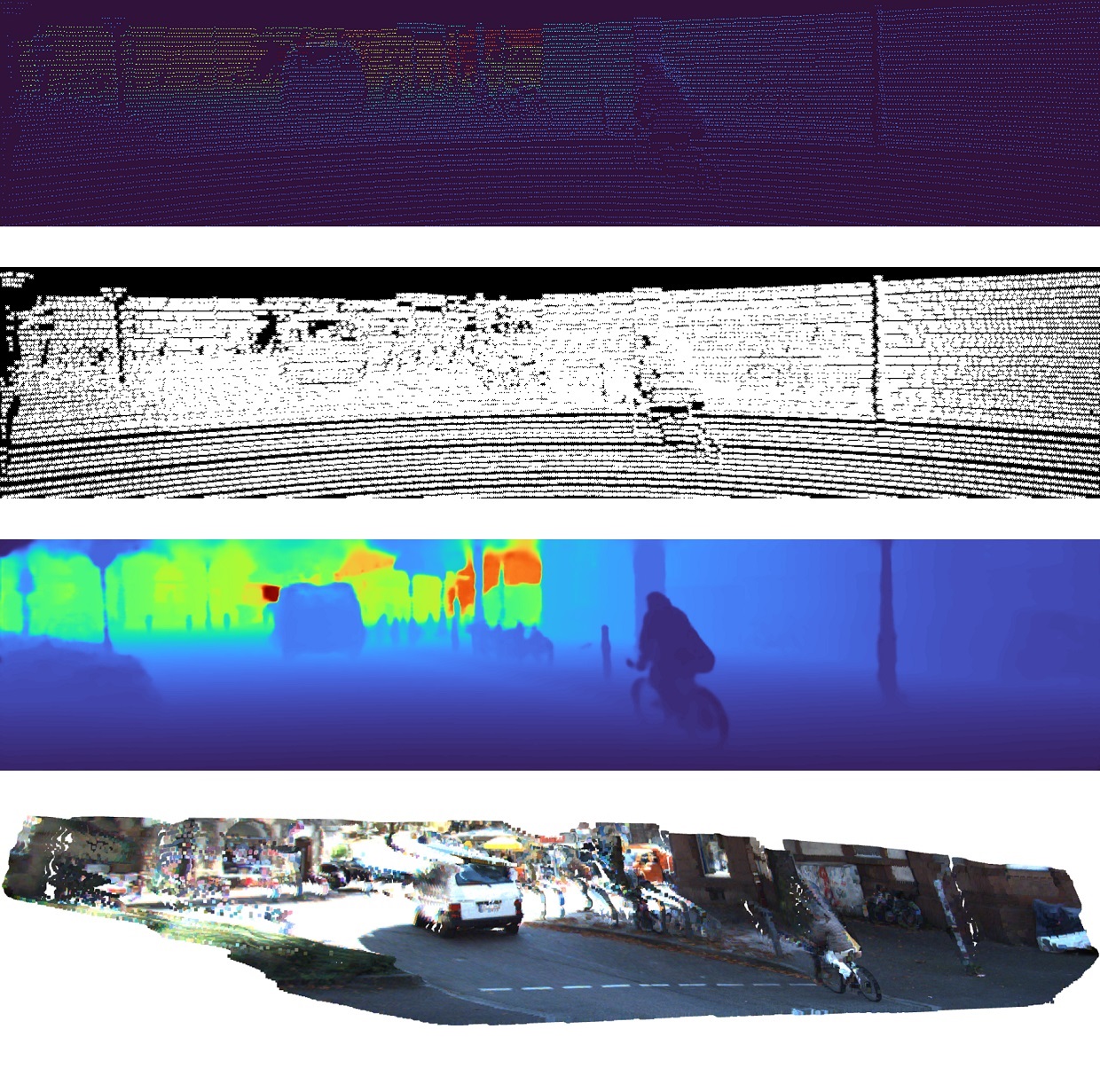}
		
		\put(50,75.5){\makebox[0pt]{\parbox[c]{200pt}{\centering \scriptsize (a) Sparse LiDAR input}}}
		\put(50,50.5){\makebox[0pt]{\parbox[c]{200pt}{\centering \scriptsize (b) Output validity mask SISL}}}
		\put(50,25.5){\makebox[0pt]{\parbox[c]{200pt}{\centering \scriptsize (c) Completed depth map}}}
		\put(50,1.5){\makebox[0pt]{\parbox[c]{200pt}{\centering \scriptsize (d) Colorized 3D point cloud}}}

	\end{overpic}
	\caption{An example of our networks depth completion (c) using KITTI data. Our novel Sparsity Invariant Switch Layer obtains a dense validity mask (b) for sparse LiDAR input (a). The colorized 3D point-cloud is shown in (d).}
	\label{pic:Teaser_Image}
\end{figure}
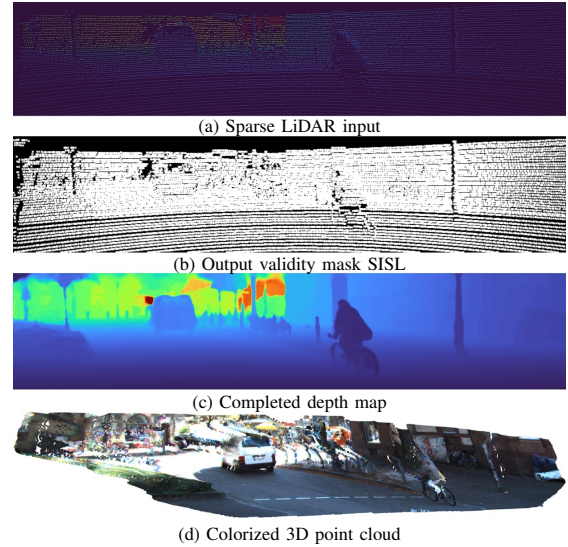

The need for sensor-driven environmental guidance is increasing. Autonomous cars, drones, or industrial automation solutions rely on a combination of different technologies to perceive their surroundings, make decisions, and overcome challenges. In the field of autonomous driving, the use of stereo cameras, radar sensors, and Light Detection And Ranging (LiDAR) sensors is common for environmental perception. Rotating LiDAR sensors create 3D depth maps, measuring distances by emitting laser pulses. The data density in the produced depth maps depends on the sensors amount of laser scan-lines. Typical LiDAR depth maps such as the one in Figure \ref{pic:Teaser_Image} have a high amount of unobserved space. Such sparse depth maps are insufficient for complex tasks such as autonomous navigation. 

The completion of sparse depth maps is an ongoing field of research. Early approaches were based on handcrafted filters and algorithms manipulating the sparse data. Current state of the art solutions rely on neural networks to complete depth maps. Neural network depth completion is split into two areas: unguided completion using only depth input, and guided completion benefiting from additional camera input. Guided networks show the most promising results, as the dense image information provides valuable features which aid in the completion of depth maps. This guidance has also been used for the benefit of scene flow estimation \cite{SceneFlow}.

We introduce a straightforward, yet effective Convolutional Neural Network (CNN) with a dual encoder-single decoder structure including skip connections, similar to \textit{U-Net} \cite{U_Net}. Our contributions are the following:
\begin{itemize}
	\item A network architecture focused on gathering dense and valid data from sparse depth maps.
	\item A novel layer to overcome sparse data, using spatially variant and content-depended dilation to gather additional neighborhood information and to reduce the spatial degradation of filters.
	\item Integrating sparsity invariant convolution into a bottleneck structure and taking into account the propagation of valid data.
\end{itemize}

\section{RELATED WORK}
\subsection{State of the Art - Depth Completion}
CNNs make up the majority of state of the art networks for depth completion, frequently utilizing encoder-decoder structures with skip connections \cite{U_Net}. Sparse input presents a challenge to the standard convolutional operation, leading to performance decreases of CNNs. As the filter moves along sparse input, its receptive field can have varying patterns and amounts of valid data, directly affecting the output value.

Uhrig \textit{et al.} \cite{uhrig2017sparsity} introduce the KITTI depth completion benchmark and propose sparsity invariant convolution (SI-convolution). SI-convolution considers only observed input through  normalization and weighing, using a binary validity mask. Similarly, binary masks have been used to filter invalid values \cite{ADNN}. The MA-bottleneck block of Yan \textit{et al.} \cite{Revisiting_Sparsity} combines SI-convolution with a residual bottleneck block \cite{ResBottleneck}, as this block aids in gradient propagation, reduces parameters and saves computational costs. Furthermore they propose the MA-fusion module, effectively combining features at decoder skip connections while reintroducing binary validity information. Jaritz \textit{et al.} \cite{Sparse_Training_Spade} reason that deep CNNs can learn to overcome sparsity without masks and focus training with varying depth map densities.

Because binary validity masks saturate in deep networks, Eldesokey \textit{et al.} \cite{Confidence_Original, Confidence_L2} use continuous confidence maps to propagate data reliability instead of validity throughout the entire network. Later networks apply confidence with surface normal guidance \cite{DeepLidar_synthdata}, with the combination of both guiding recurrent refinement \cite{PWP}. These solutions train their normal prediction with synthetic data \cite{DeepLidar_synthdata} or based on principal component analysis \cite{PWP}. Others use confidence to fuse image and depth features, giving more weight to the modality with less uncertainty \cite{Sparse_Uncertain,CrossGuidance,GraphConv}.

\textit{SPN} \cite{SPN_Affinity} proposes learned affinity matrices from guidance input, followed by spatial propagation for the refinement of segmentation masks. Affinity based refinement also benefits  completed depth maps \cite{CSPN}, strengthening object alignment and reducing blurry results. Some methods use refinement conjointly with confidence maps \cite{DefSPN,CSPN++,PE_Net}. However, spatial propagation only works within a fixed local neighborhood. Adding deformable convolution \cite{DefSPN}, learned non-local neighbor information \cite{NL_spatialprop}, or trainable parameters into the recurrent refinement process \cite{CSPN++} further improves accuracy. Schuster \textit{et al.} \cite{SSGP_Wasenmueller} propose an image guided, sparsity-aware, convolutional module, with subsequent spatial propagation refinement. Sparsity-awareness is achieved through a binary validity mask.

Multi-scale information improves the capability of networks to overcome differently sized or deformed input. Various networks integrate Spatial Pyramid Pooling (SPP) \cite{SPP-KaimingHe} for depth completion \cite{Revisiting_Sparsity,CSPN,CSPN++,DFUSE}. Atrous Spatial Pyramid Pooling (ASPP) \cite{DeepLabv3} has been studied at the end of an encoder \cite{CSPN} or within residual blocks \cite{CrossGuidance}. Li \textit{et al.} \cite{MSG-CHN} combine multiple networks, each using different resolutions of sparse input. Re-scaled input is also used for the fusion of 2D and 3D information \cite{2D_3D}.

\subsection{State of the Art - Dilated Convolution}
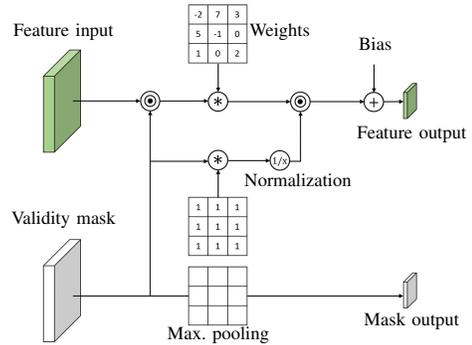
\begin{figure}[t]
	\centering
	\begin{overpic}[width=0.7\columnwidth, ,tics=5]
		{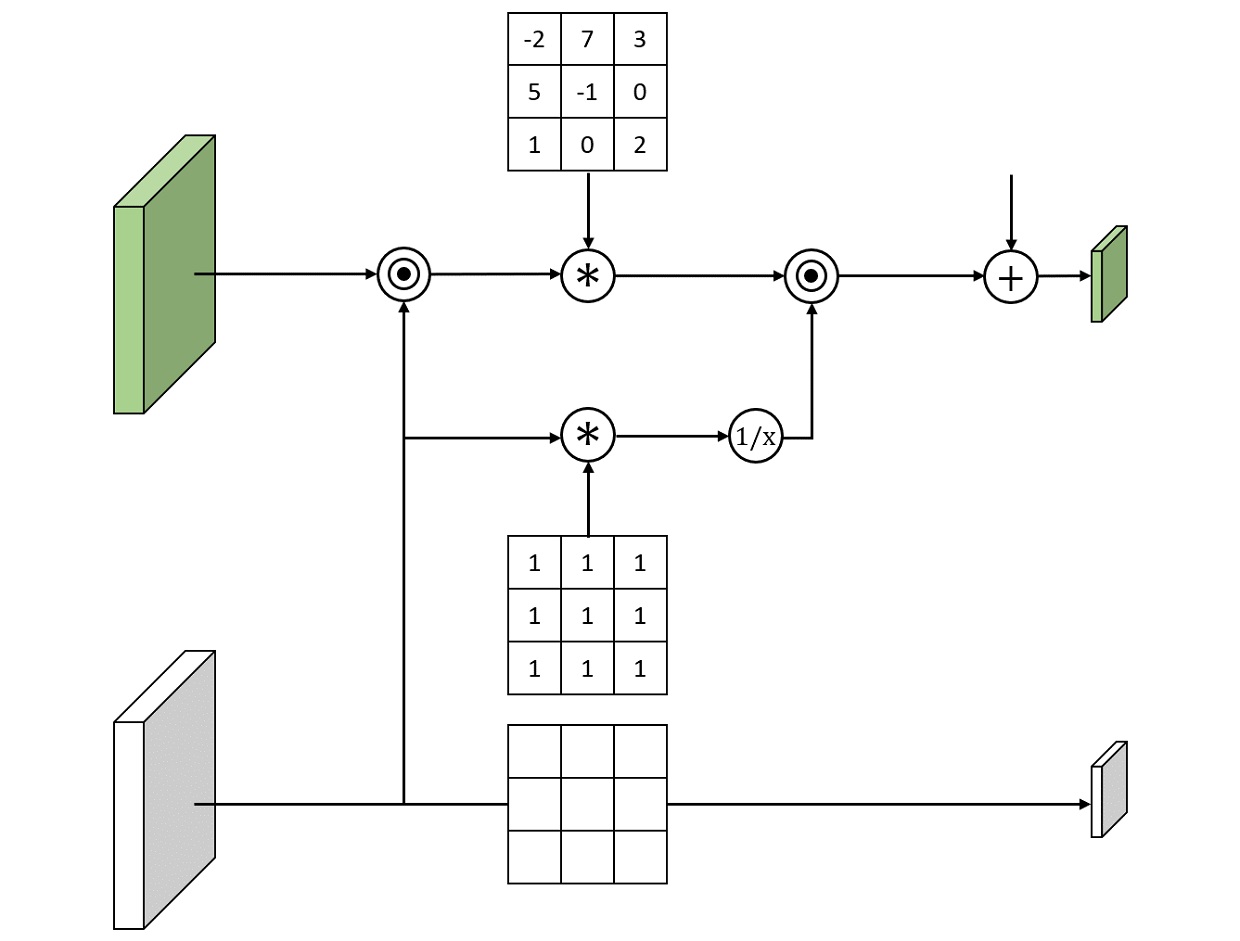}
		\put(13.5,67.5){\makebox[0pt]{\parbox[c]{200pt}{\centering \scriptsize Feature input}}}
		\put(13.5,26.25){\makebox[0pt]{\parbox[c]{200pt}{\centering \scriptsize Validity mask}}}
		\put(61,67.5){\makebox[0pt]{\parbox[c]{200pt}{\centering \scriptsize Weights}}}
		\put(65,35){\makebox[0pt]{\parbox[c]{200pt}{\centering \scriptsize Normalization}}}
		\put(47.5,1){\makebox[0pt]{\parbox[c]{200pt}{\centering \scriptsize Max. pooling}}}
		\put(82,65){\makebox[0pt]{\parbox[c]{200pt}{\centering \scriptsize Bias}}}
		\put(90,45){\makebox[0pt]{\parbox[c]{200pt}{\centering \scriptsize Feature output}}}
		\put(90,4){\makebox[0pt]{\parbox[c]{200pt}{\centering \scriptsize Mask output}}}
	\end{overpic}
	\caption{SI-Convolution. Here $\odot$ represents element-wise multiplication and $*$ convolution. Adapted from Uhrig \textit{et al.} \cite{uhrig2017sparsity}.} \label{pic:Original_SI_Convolution}
\end{figure}

Dilated convolution \cite{Dilated_origin} learns features at different scale, without changing the spatial size of feature maps, adding parameters, or computational burden. However, the filters effectiveness depends on the dilation rate. The "holes" in the dilated filter also produce a gridding effect, skipping over direct neighborhood information when calculating an output value \cite{Dilation_Gridding}. The \textit{DeepLab} series \cite{DeepLabv1,DeepLabv2,DeepLabv3} and \textit{PSP} \cite{PSP_net} integrate dilated convolution to replace deeper pooling layers within a network, retaining spatial dimensions. \textit{DeepLab} also introduced ASPP, using parallel dilated convolution to gather multi-scale information. The Stacked Dilated Convolution (SDC) layer uses different dilation rates in parallel to make up an entire network \cite{SDC}. Qiao \textit{et al.} \cite{DetectoRS} introduce the Switchable Atrous Convolution (SAC) layer, utilizing a single attention map to spatially combine the output from different dilation rates. Likewise, Li \textit{et al.} \cite{SK_Net} use attention within their layer to combine feature maps of different dilation rates in the channel dimension.

\section{METHODS}
The aforementioned depth completion methods broadly study depth refinement, predominantly using learned affinity maps and spatial propagation. Confidence propagation has also been applied extensively, in some networks combined with refinement. Current state of the art solutions reveal two research gaps. To the best of our knowledge, there has been no implementation of a SI-convolutional network specifically focusing on the amount of valid pixels gathered from sparse data. Moreover, while multi-scale information is common, there has been limited research using dilated convolution for the benefit of gathering additional data from depth maps.

Based on this review, we formulate two primary research goals: increasing valid information gathered from sparse data by focusing on dense validity masks, and further exploring dilated convolution in the context of depth completion. 

\subsection{Sparsity Invariant Convolution}

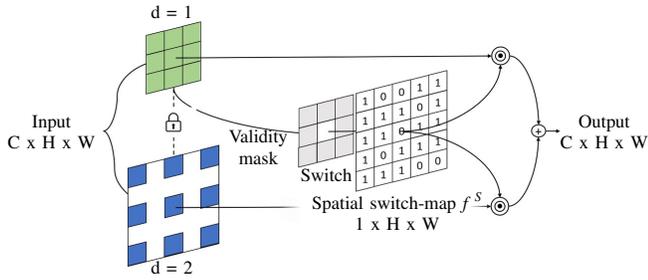
\begin{figure}[t]
	\centering
	\begin{overpic}[width=0.95\columnwidth, ,tics=5]
		{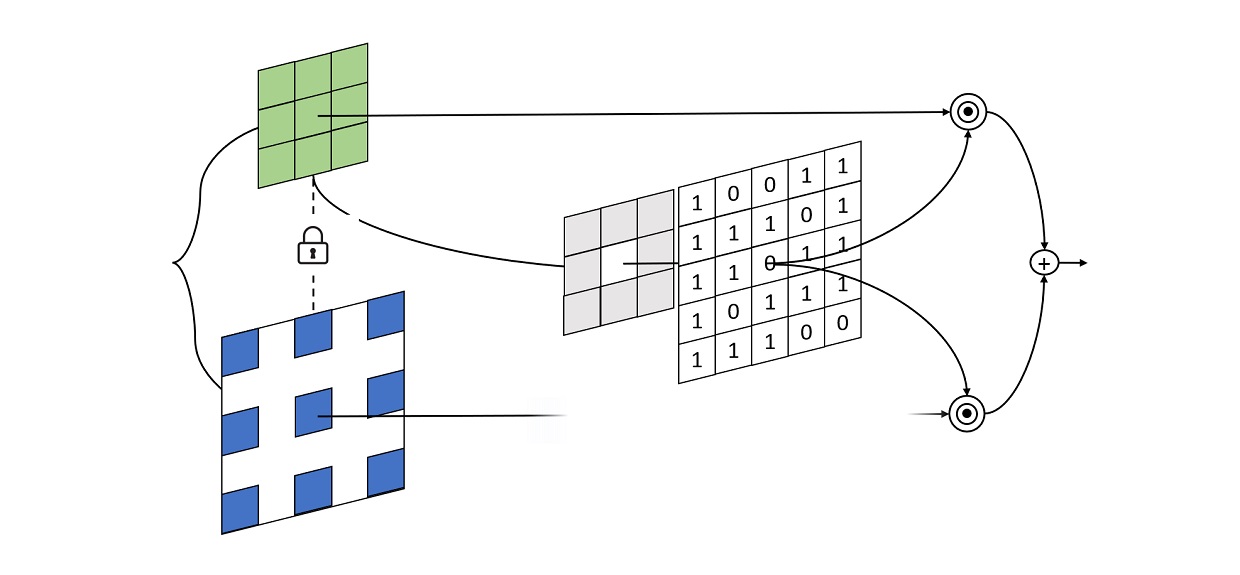}
		\put(5.5,24){\makebox[0pt]{\parbox[c]{200pt}{\centering \scriptsize Input\\C x H x W}}}
		\put(95,24){\makebox[0pt]{\parbox[c]{200pt}{\centering \scriptsize Output\\C x H x W}}}
		\put(50,17){\makebox[0pt]{\parbox[c]{80pt}{\centering \scriptsize Switch}}}
		\put(61.5,11.2){\makebox[0pt]{\parbox[c]{200pt}{\centering \scriptsize Spatial switch-map \textit{f\textsuperscript{\, S}}\\1 x H x W}}}
		\put(39,21.5){\makebox[0pt]{\parbox[c]{200pt}{\centering \scriptsize Validity\\mask}}}
		\put(25,43.5){\makebox[0pt]{\parbox[c]{200pt}{\centering \scriptsize d = 1}}}
		\put(25,2){\makebox[0pt]{\parbox[c]{200pt}{\centering \scriptsize d = 2}}}
	\end{overpic}
	\caption{Our Sparsity Invariant Switch Layer (SISL), here with a content-depended dilation rate of $d=2$, enables a network to gather additional information from sparse data. The lock represents optional weight sharing allowing SISL to be seamlessly integrated into any SI-convolution. Element-wise multiplication is represented by $\odot$.}
	\label{pic:Switchable Input Layer}
\end{figure}

Because of its fundamental relevance to our developments, it is necessary to review Sparsity Invariant Convolution \cite{uhrig2017sparsity}, depicted in Figure \ref{pic:Original_SI_Convolution}. SI-Convolution aims to improve convolutional operation on sparse data. Depth maps are projected into 2.5D space and the location of observed pixels are encoded in a binary validity mask. This mask is used by SI-convolution to weight and normalize the elements within the filter. The output $f_{u,v}(x,o)$ of SI-convolution is given analogous to Uhrig \textit{et al.} \cite{uhrig2017sparsity} by
\begin{equation} \label{eq:Sparsity_Invariant_Convolution}
	f_{u,v}(x,o)=\frac{\sum_{i,j=-k}^{k} o_{u+i\cdot d,v+j\cdot d}x_{u+i\cdot d,v+j\cdot d}w_{i,j}}{\sum_{i,j=-k}^{k} o_{u+i\cdot d,v+j\cdot d}+\epsilon}+b \text{ ,}
\end{equation}
 with the input tensor $x$ and its corresponding binary validity mask $o$ (both zero padded), the convolutional weights $w$, dilation rate $d$, and the optional bias $b$. The kernel size is (2$k$+1)\textsuperscript{2}. A small term $\epsilon > 0$ is added to the denominator to prevent division by zero. To track the validity-state of the output, SI-convolution propagates the validity mask through max pooling:
\begin{equation} \label{eq:Mask Propagation}
	f_{u,v}^{o}(o)=\max_{i,j=-k,...,k}o_{u+i\cdot d,v+j\cdot d} \text{ .}
\end{equation}
With subsequent propagation, the validity mask becomes denser.

\subsection{Sparsity Invariant Switch Layer (SISL)}
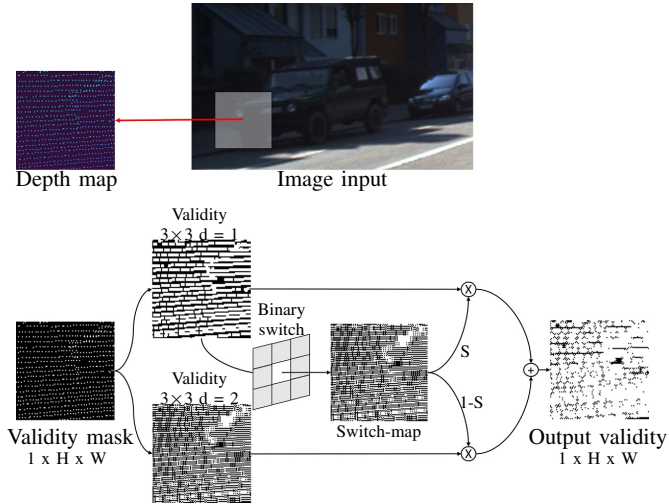
\begin{figure}[t]
	\centering
	\resizebox{\columnwidth}{!}{
		\begin{overpic}[width=0.50\textwidth, ,tics=5]{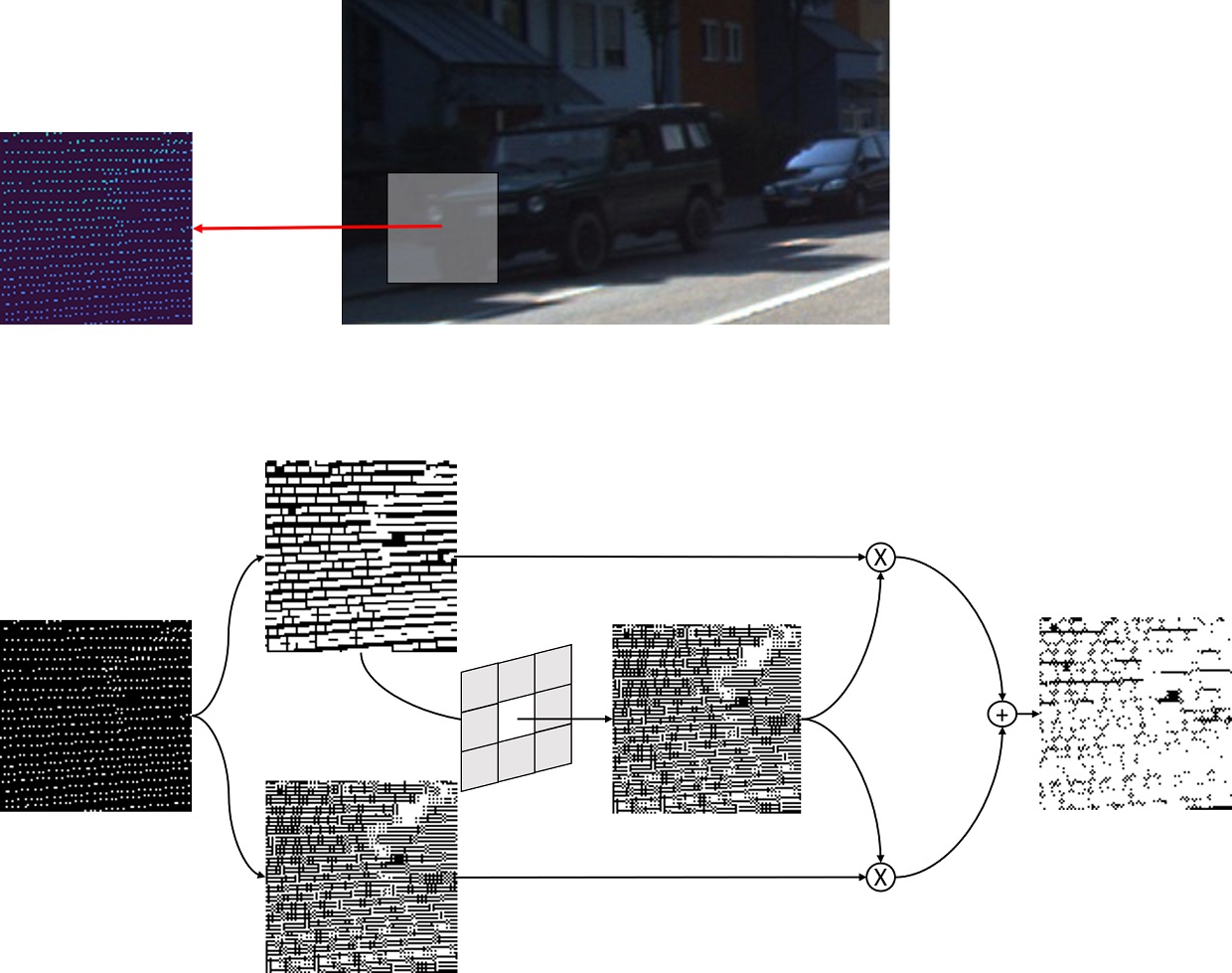}
			\put(50,50){\makebox[0pt]{\parbox[c]{100pt}{\centering \small Image input}}}
			\put(8,50){\makebox[0pt]{\parbox[c]{100pt}{\centering \small Depth map}}}
			\put(8,8){\makebox[0pt]{\parbox[c]{100pt}{\centering \small ~Validity mask\\ \scriptsize 1 x H x W}}}
			\put(29,43){\makebox[0pt]{\parbox[c]{100pt}{\centering \scriptsize Validity\\3$\times$3 d = 1}}}
			\put(29,17){\makebox[0pt]{\parbox[c]{100pt}{\centering \scriptsize Validity\\3$\times$3 d = 2}}}
			\put(42,28){\makebox[0pt]{\parbox[c]{100pt}{\centering \scriptsize Binary\\switch}}}
			\put(57.35,10){\makebox[0pt]{\parbox[c]{100pt}{\centering \scriptsize Switch-map}}}
			\put(92,8){\makebox[0pt]{\parbox[c]{100pt}{\centering \small Output validity\\ \scriptsize 1 x H x W}}}
			\put(72,15){\makebox[0pt]{\parbox[c]{100pt}{\centering \scriptsize 1-S}}}
			\put(71,23){\makebox[0pt]{\parbox[c]{100pt}{\centering \scriptsize S}}}
		\end{overpic}
	}
	\caption{SISL has dense output features through content-depended dilation. White pixels in the validity masks depict where valid data is observed. The black spaces in the switch-map show the spatial location where the dilated convolution is beneficial. The shown example is from SISL within the first layer of the network using KITTI data. In these cropped masks, the output validity of a 3$\times$3 SI-convolution is 58.46\%, while the validity of SISL is 87.84\%.}
	\label{pic:Binary_Switch}
\end{figure}

In 2.5D sparse depth maps, there are spatial areas where a convolution observes few or no valid pixels, degrading its effectiveness. In such instances, the output does not consider neighboring information or is invalid. This issue especially affects early convolution layers. SI-convolution focuses on the propagation of valid information, but not on increasing the amount of valid input. For this reason, learned feature maps continue to exhibit sparsity and scan-line patterns, until the validity mask saturates through propagation.

Thus, we propose the \textit{Sparsity Invariant Switch Layer} (SISL, see Figure \ref{pic:Switchable Input Layer}), enabling a spatially- and content-dependent increase of dilation rate, using a binary switch. This allows SISL to gather additional information and reduce the spatial degradation of filters.  We specifically choose dilated convolution over a larger receptive field for its distinct benefits. Dilated convolution requires less parameters and by extent generalizes better, saves computational cost, and re-samples its input without changing the output dimension. Moreover, dilated convolution enables weight sharing between parallel layers, which allows SISL to be integrated seamlessly into any existing SI-convolution.

The binary switch function considers the content in a convolution filter by observing the binary validity mask. We use a switch, since the validity mask encodes the precise location of information, compared to learning this information through attention. In the case of a 3$\times$3 convolution, the switch increases the dilation rate only if all eight outside pixels are empty. We specifically choose eight empty pixels, so neighbor information is not skipped. This spatially variant dilation rate alleviates the gridding effect, because data is only resampled where no information in the receptive field is excluded. The content in the center pixel remains the same regardless of dilation and it is not considered by the switch. The proposed binary switch function can be described by
\begin{equation} \label{eq: Binary Switch}
	f_{u,v}^{S}(o)=\min \left(\sum_{i,j=-k}^{k}o_{u+i,v+j}-o_{u,v} ,1\right)
\end{equation}
with the undilated kernel size = (2\textit{k}+1)\textsuperscript{2}, the validity mask \textit{o}, and the resulting switch-map $f_{u,v}^{S}(o)$. The output of SISL is
\begin{equation} \label{eq: Switchable Input Layer}
	\text{output} = f^{S}(o)f^{d = 1}(x,o)+\left(1-f^{S}(o)\right)f^{d = 2}(x,o)
\end{equation}
with $f^{d=1}(x,o)$ and $f^{d=2}(x,o)$ as the output of SI-convolution (based on Equation \ref{eq:Sparsity_Invariant_Convolution}) with dilation rates $d$. The switch-map is also applied for validity mask propagation
\begin{equation} \label{eq: Switchable Input Layer}
	\text{mask} = f^{S}(o)f^{o, d=1}(o)+\left(1-f^{S}(o)\right)f^{o,d=2}(o)
\end{equation}
with $f^{o,d=1}(o)$ and  $f^{o,d=2}(o)$ as the output validity mask of the convolutions, according to Equation \ref{eq:Mask Propagation}. The effect of SISL on the validity mask is shown in Figure \ref{pic:Binary_Switch}.

\subsection{SI-Residual Bottleneck}
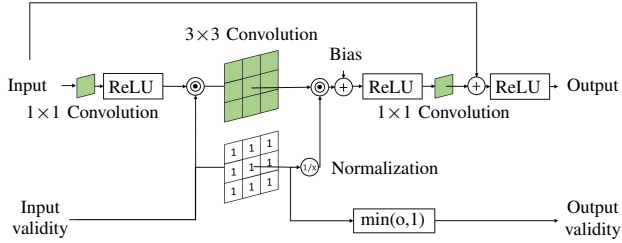
\begin{figure}[t]
	\centering
	\begin{overpic}[width=\columnwidth, ,tics=5]
		{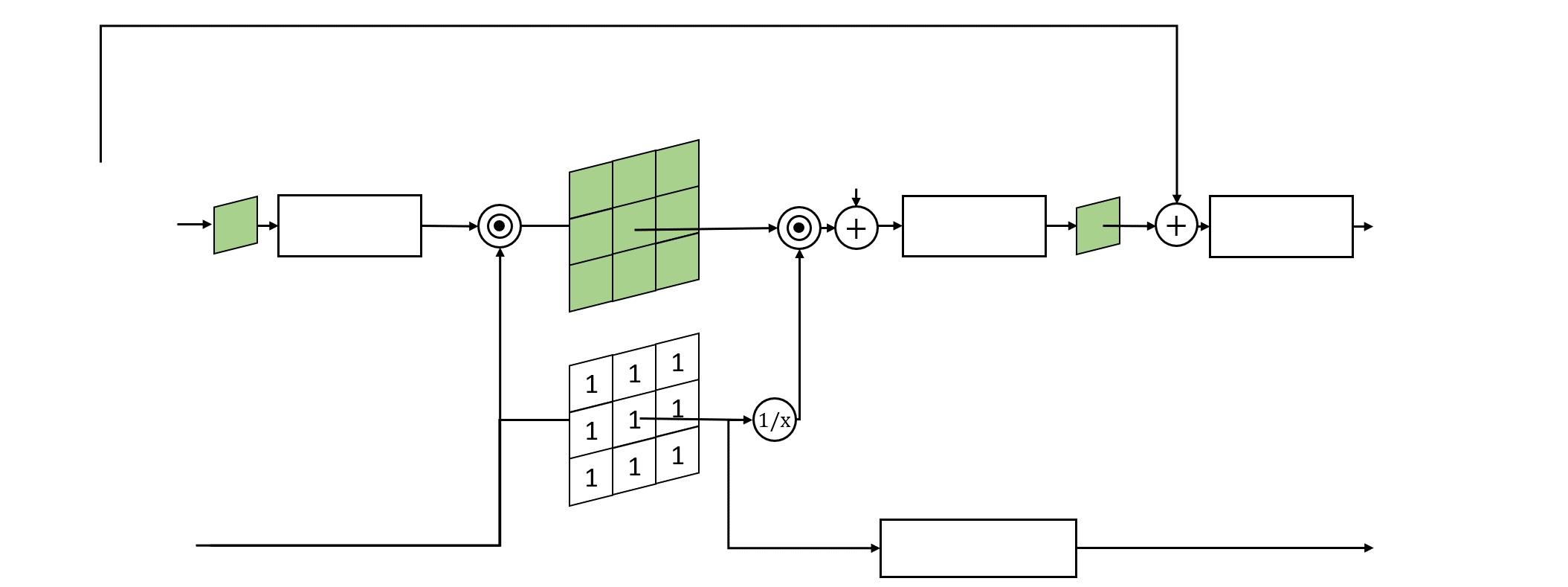}
		\put(22,22){\makebox[0pt]{\parbox[c]{100pt}{\centering \scriptsize ReLU}}}
		\put(81.5,22){\makebox[0pt]{\parbox[c]{100pt}{\centering \scriptsize ReLU}}}
		\put(62,22){\makebox[0pt]{\parbox[c]{100pt}{\centering \scriptsize ReLU}}}
		\put(6,22){\makebox[0pt]{\parbox[c]{100pt}{\centering \scriptsize Input}}}
		\put(93,22){\makebox[0pt]{\parbox[c]{100pt}{\centering \scriptsize Output}}}
		\put(62.25,1.4){\makebox[0pt]{\parbox[c]{100pt}{\centering \scriptsize min(o,1)}}}
		\put(8,1.5){\makebox[0pt]{\parbox[c]{100pt}{\centering \scriptsize Input\\validity}}}
		\put(93,1.5){\makebox[0pt]{\parbox[c]{100pt}{\centering \scriptsize Output\\validity}}}
		\put(55,27){\makebox[0pt]{\parbox[c]{100pt}{\centering \scriptsize Bias}}}
		\put(40.5,30){\makebox[0pt]{\parbox[c]{100pt}{\centering \scriptsize 3$\times$3 Convolution}}}
		\put(16,18){\makebox[0pt]{\parbox[c]{100pt}{\centering \scriptsize 1$\times$1 Convolution}}}
		\put(70,18){\makebox[0pt]{\parbox[c]{100pt}{\centering \scriptsize 1$\times$1 Convolution}}}
		\put(61,9.5){\makebox[0pt]{\parbox[c]{100pt}{\centering \scriptsize Normalization}}}
	\end{overpic}
	\caption{The Sparsity Invariant Residual Bottleneck, shown for a 3$\times$3 convolution. Here $\odot$ represents element-wise multiplication.}\label{pic:Bottleneck_v2}
	\label{pic:Kap3-Redesigned ResBottleneck}
\end{figure}

Furthermore, we propose the SI-Residual Bottleneck, inspired by He \textit{et al.} \cite{ResBottleneck} and Yan \textit{et al.} \cite{Revisiting_Sparsity}. However, Yan \textit{et al.} \cite{Revisiting_Sparsity} limit mask propagation to down-sampling layers. Sequential use of the same validity mask zeroes valid output at the next layer. In contrast, our bottleneck considers mask propagation, retaining valid information in the network and adding robustness to varying sparsity.

A bottleneck structure allows for deeper networks, reducing parameters and computational costs. Analogous to He \textit{et al.} \cite{ResBottleneck}, our SI-Residual Bottleneck (Figure \ref{pic:Kap3-Redesigned ResBottleneck}) uses three consecutive layers: a 1$\times$1 convolution reduces the input channels (bottleneck width), followed by e.g. a 3$\times$3 SI-convolution, before another 1$\times$1 convolution resizes the features to the desired output channels. Afterwards, the residual is added. If used for expansion, the residual of the bottleneck is resized by another 1$\times$1 convolution. For validity propagation, we use the convoluted binary mask which is utilized for weighing and normalization in the SI-operation, and limit it to a maximum value of one. The result is identical to Equation \ref{eq:Mask Propagation}, however zero padding is not necessary. SI-convolution is not applied in the 1$\times$1 convolutions, as these cannot observe neighboring information within their receptive field.

\subsection{Dense Validity Mask Network}
\label{Kap:Final_NN}
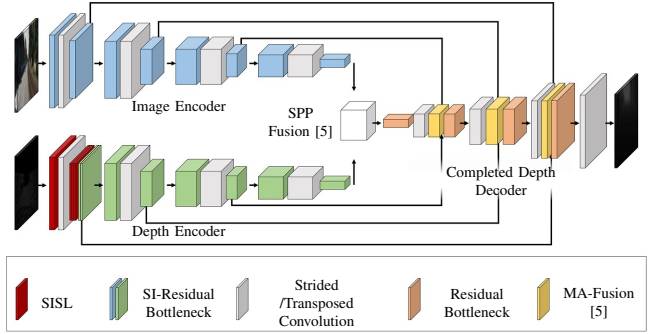
\begin{figure}[t]
	\centering
	\resizebox{\columnwidth}{!}{
		\begin{overpic}[width=\textwidth, ,tics=5]
			{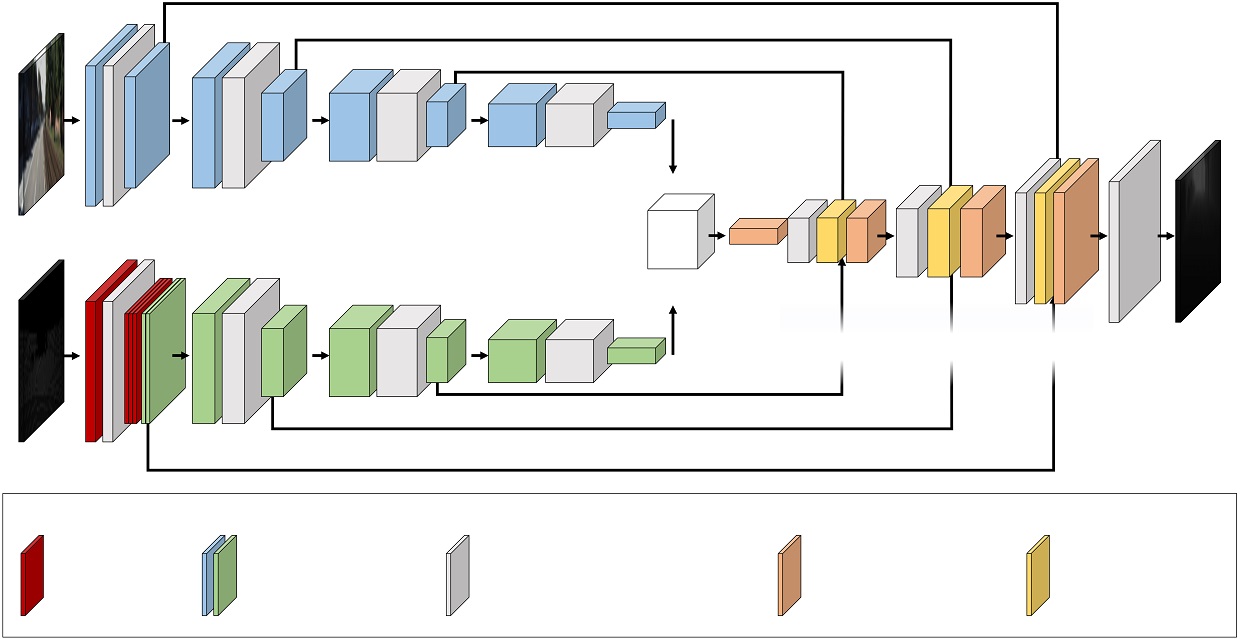}
			\put(8,4){\makebox[0pt]{\parbox[c]{100pt}{\centering \large SISL}}}
			\put(27,4){\makebox[0pt]{\parbox[c]{100pt}{\centering \large SI-Residual\\Bottleneck}}}
			\put(48.5,4){\makebox[0pt]{\parbox[c]{100pt}{\centering \large Strided \\ /Transposed Convolution}}}
			\put(73.5,4){\makebox[0pt]{\parbox[c]{100pt}{\centering \large Residual\\ Bottleneck}}}
			\put(92.5,4){\makebox[0pt]{\parbox[c]{100pt}{\centering \large MA-Fusion\\ \cite{Revisiting_Sparsity}}}}
			\put(46,32){\makebox[0pt]{\parbox[c]{100pt}{\centering \large SPP\\Fusion \cite{Revisiting_Sparsity}}}}
			
			\put(27,15){\makebox[0pt]{\parbox[c]{100pt}{\centering \large Depth Encoder}}}
			\put(27,34.5){\makebox[0pt]{\parbox[c]{100pt}{\centering \large Image Encoder}}}
			\put(77.25,23){\makebox[0pt]{\parbox[c]{100pt}{\centering \large Completed Depth Decoder}}}
		\end{overpic}
	}
	\caption{Dense Validity Mask Network architecture. The first depth encoder stage is depicted in detail, highlighting the use of Sparsity Invariant Switch Layers (SISL). Within this encoder stage, our SISL replaces SI-convolution for the first four SI-Residual Bottlenecks.}
	\label{pic:Kap4:Final_Model}
\end{figure}

We name our final neural network \textit{Dense Validity Mask Network} (DVMN), due to its focus on a dense validity mask. DVMN utilizes both SISL and the "plain" SI-Residual Bottleneck, without weight sharing. The network is built as a dual encoder, single decoder structure, with skip connections. Its architecture is depicted in Figure \ref{pic:Kap4:Final_Model}.

\subsubsection{Depth and Image Encoders}
Both the depth and image encoders consist of four stages. Each stage expands the channels by $C\cdot$stage using a SI-Residual Bottleneck, then down-samples using strided SI-convolution, followed by five subsequent SI-Residual Bottlenecks. This results in an encoder output of 128 feature maps at $\frac{1}{16}$ height and $\frac{1}{16}$ width. Due to the dense input, SI-convolution acts as a regular convolution in the image encoder. Both encoders are identical, but differ in one aspect: SISL is used in the first four bottlenecks of the first depth encoder stage.

The output of both encoders are fused with added multi-scale SPP context analogous to Yan \textit{et al.} \cite{Revisiting_Sparsity}, reducing features at a ratio of 2:1.

\subsubsection{Decoder}
The decoder consists of four stages. The first three stages are identical, using transposed convolution to up-sample feature maps, and reducing the channel amount to  4$C$ - $C \cdot$stage. This is followed by skip connections using the MA-Fusion block \cite{Revisiting_Sparsity} and a regular residual bottleneck. The decoder includes batch-normalization. The fourth encoder stage is a single transposed convolution, generating the completed depth map of original input dimensions.

\begin{table}[t]
	\caption{Ablation study comparing designs of our SI-Residual Bottleneck.}
	\resizebox{\columnwidth}{!}{
		\begin{tabular}{ccccc|cc}
			\hline
			\multicolumn{1}{c|}{\multirow{2}{*}{\textbf{Version}}} & \multicolumn{1}{c|}{\multirow{2}{*}{\textbf{Bottleneck}}} & \multicolumn{1}{c|}{\multirow{2}{*}{\textbf{\begin{tabular}[c]{@{}c@{}}Mask\\ Prop.\end{tabular}}}} & \multicolumn{1}{c|}{\multirow{2}{*}{\textbf{\begin{tabular}[c]{@{}c@{}}Pre-\\ activation\end{tabular}}}} & \multirow{2}{*}{\textbf{\begin{tabular}[c]{@{}c@{}}Pre-\\ addition\end{tabular}}} & \multicolumn{2}{c|}{\textbf{Error Metric}}                                                                                                      \\
			\multicolumn{1}{c|}{}                                  & \multicolumn{1}{c|}{}                                     & \multicolumn{1}{c|}{}                                                                                     & \multicolumn{1}{c|}{}                                                                                    &                                                                                   & \multicolumn{1}{c|}{\textbf{\begin{tabular}[c]{@{}c@{}}RMSE\\ (mm)\end{tabular}}} & \textbf{\begin{tabular}[c]{@{}c@{}}MAE\\ (mm)\end{tabular}} \\ \hline
			\multicolumn{2}{c}{MA-Bottleneck \cite{Revisiting_Sparsity}}                                                                              &                                                                                                           &                                                                                                          &                                                                                   & 705.9                                                                             & 208.1                                                       \\ \hline
			v1                                                     & ours                                    & \checkmark                                                                                                       &                                                                                                          &                                                                                   & \textbf{688.7}                                                                    & 204.1                                                       \\
			v2                                                     & ours                                    & \checkmark                                                                                                       & \checkmark                                                                                                      &                                                                                   & 769.0                                                                             & 215.8                                                       \\
			v3                                                     & ours                                    & \checkmark                                                                                                       &                                                                                                          & \checkmark                                                                               & 774.0                                                                             & 212.9                                                       \\ \hline
		\end{tabular}
	}
	\label{Tb: Ablation-SI-resbn}
	
	\caption{Ablation study of SISL. E1 signifies the switch location within the first stage of the depth encoder. The dilation rate is d. v1 is chosen for the neural network.}
	\resizebox{\columnwidth}{!}{
		\begin{tabular}{ccccccc}
			\hline
			\multicolumn{1}{c|}{\multirow{2}{*}{\textbf{Version}}} & \multicolumn{1}{c|}{\multirow{2}{*}{\textbf{Switch Design}}}                        & \multicolumn{1}{c|}{\multirow{2}{*}{\textbf{\begin{tabular}[c]{@{}c@{}}Mask\\ Prop.\end{tabular}}}} & \multicolumn{1}{c|}{\multirow{2}{*}{\textbf{\begin{tabular}[c]{@{}c@{}}Weight\\ Sharing\end{tabular}}}} & \multicolumn{1}{c|}{\multirow{2}{*}{\textbf{d}}} & \multicolumn{2}{c}{\textbf{Error Metric}}                                                                                 \\
			\multicolumn{1}{c|}{}                                  & \multicolumn{1}{c|}{}                                                               & \multicolumn{1}{c|}{}                                                                                     & \multicolumn{1}{c|}{}                                                                                   & \multicolumn{1}{c|}{}                                   & \textbf{\begin{tabular}[c]{@{}c@{}}RMSE\\ (mm)\end{tabular}} & \textbf{\begin{tabular}[c]{@{}c@{}}MAE\\ (mm)\end{tabular}} \\ \hline
			v1                                                     & Binary                                                                              & \checkmark                                                                                                       &                                                                                                         & 2                                                       & 687.4                                             & 202.8                                                       \\
			v2                                                     & Binary                                                                              & \checkmark                                                                                                       &                                                                                                         & 3                                                       & 701.5                                                        & 209.1                                                       \\
			v3                                                     & Binary                                                                              & \checkmark                                                                                                       &                                                                                                         & 4                                                       & 702.6                                                        & 209.4                                                       \\ \hline
			v4                                                     & Binary                                                                              & \checkmark                                                                                                       & \checkmark                                                                                                     & 2                                                       & 696.4                                                        & 205.2                                                       \\
			v5                                                     & Binary                                                                              &                                                                                                           &                                                                                                         & 2                                                       & 697.1                                                        & 207.5                                                       \\
			v6                                                     & None                                                                                & \checkmark                                                                                                       &                                                                                                         & 2                                                       & 691.4                                                        & 206.2                                                       \\ \hline
			-                                                      & Binary E1 + Attention                                                            &                                                                                                           & \checkmark                                                                                                     & 2                                                       & 696.3                                                        & 210.3                                                       \\
			-                                                      & \begin{tabular}[c]{@{}c@{}}Binary E1 + Attention\\ + Global Context\end{tabular} & \checkmark                                                                                                       &                                                                                                         & 2                                                       & 690.7                                                        & 202.1                                                       \\
			-                                                      & Binary E1 + Attention                                                            & \checkmark                                                                                                       &                                                                                                         & 2                                                       & \textbf{682.4}                                                        & 201.8                                                       \\ \hline
		\end{tabular}
	}
	\label{Tb: Ablation-SISL}	
\end{table}

\section{Evaluation}
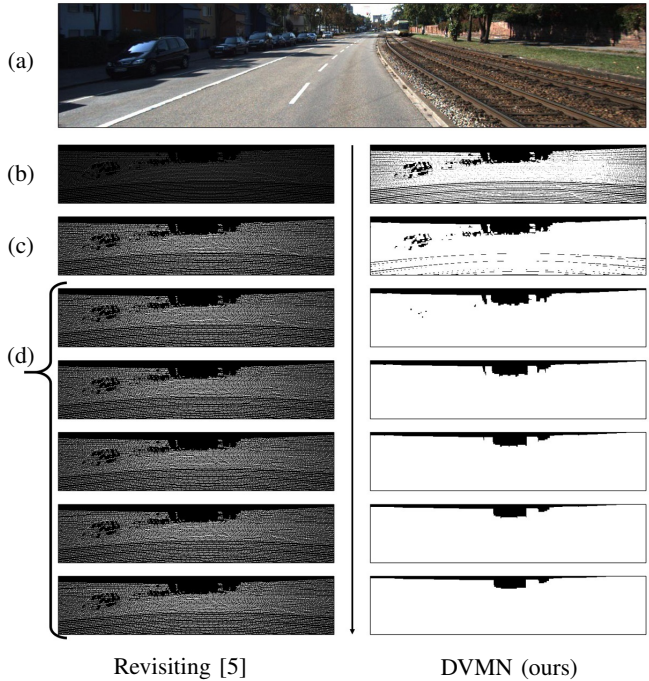
\begin{figure}[t]
	\centering
	\begin{overpic}[width=\columnwidth, ,tics=5]
		{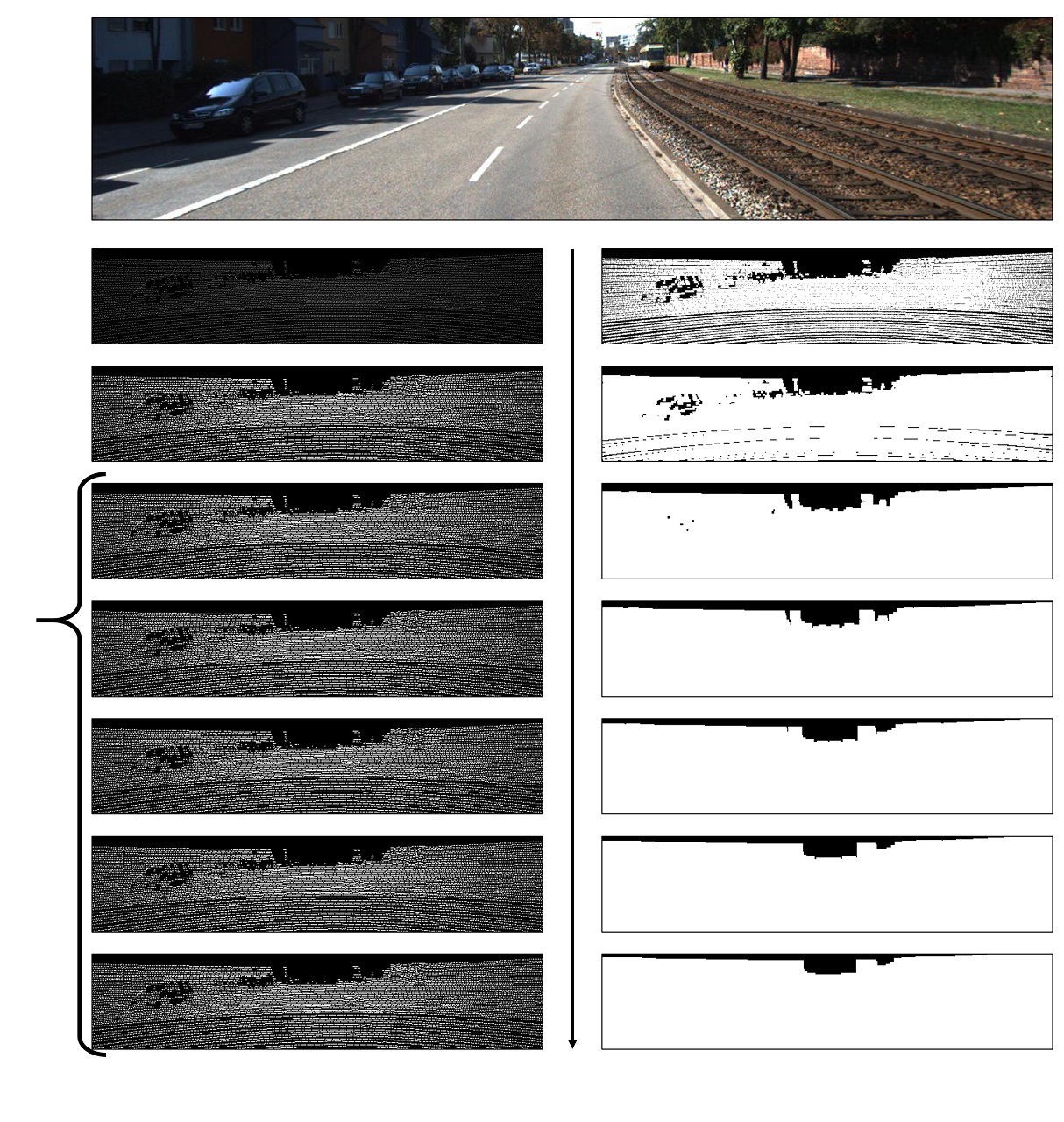}
		\put(2.75,89){\makebox[0pt]{\parbox[c]{100pt}{\centering \small (a)}}}
		\put(2.75,72.5){\makebox[0pt]{\parbox[c]{100pt}{\centering \small (b)}}}
		\put(2.75,62){\makebox[0pt]{\parbox[c]{100pt}{\centering \small (c)}}}
		\put(2.75,46){\makebox[0pt]{\parbox[c]{100pt}{\centering \small (d)}}}
		
		\put(26,0.5){\makebox[0pt]{\parbox[c]{200pt}{\centering \small Revisiting \cite{Revisiting_Sparsity}}}}
		\put(74,0.5){\makebox[0pt]{\parbox[c]{200pt}{\centering \small DVMN (ours)}}}
		
	\end{overpic}
	\caption{The output validity of the first seven layers of DVMN compared to \textit{Revisiting} \cite{Revisiting_Sparsity}. White pixels represent a valid output. The input layers (b) are downsized for this depiction, in the case of DVMN this is a SISL, followed by the down-sampling layers (c). DVMN uses three SISL and two SI-Residual Bottlenecks in (d), compared to five layers of \textit{Revisiting}. The input image is depicted in (a). Since these validity masks are applied to learned feature maps in SI-convolution, our focus on a saturated validity mask reduces the scan-line pattern and sparsity in learned features.
	}
	\label{pic:Kap4:DVMN_vs_Rev}
\end{figure}
We perform the ablation and evaluation of our methods using the KITTI depth completion dataset \cite{uhrig2017sparsity}. The dataset contains sparse depth maps and aligned RGB-images, with splits of approximately 86K data-pairs for training, 1k for validation, and 1k for testing. Due to the LiDAR sensors vertical field of view, the top areas of the 2.5D depth maps do not contain measured nor ground-truth values. The sparse depth maps have approximately 5\% observed data. For our ablation study we use the Root Mean Square Error (RMSE) and mean absolute error (MAE) error metrics. The benchmark uses the RMSE for ranking.

For ablation we use the ADAM optimizer without weight decay, and the plateau learning rate schedule using a reduction factor of 0.5. The training duration is 50 epochs with an initial learning rate of 0.001 and a batch size of 4. For data augmentation we use horizontal and vertical axis flipping, random rotation, and added Gaussian noise. The loss function uses the mean squared error and adds smoothness loss scaled by the hyper-parameter $\lambda$ = 0.1, to reduce gridded depth values \cite{Smooth_Loss}. We implement a dual encoder-single decoder structure with four stages and skip connections. Each encoder stage increases the channel dimension by $C$ = 32. We use bottleneck widths of 0.5. The final network is described in \ref{Kap:Final_NN}. 

\subsection{Ablation Studies}
\subsubsection{Sparsity Invariant Switch Layer}
We remove individual components of SISL to verify their impact. We also study the layer with changed dilation rates and an attention switch. The results are shown in Table \ref{Tb: Ablation-SISL}. 

In the first study we construct a depth decoder entirely from SISL (v1), excluding the down-sampling layers. The validity mask saturates with subsequent propagation, reducing the switch-activation. Through a mask analysis we determine that SISL impacts the first encoder stage. Beyond this initial stage, SISL continues to increases dilation in large areas of sparsity without a ground-truth, such as the top of depth maps or areas with measurement errors. For further ablation, we use SISL within the first depth encoder stage. Introducing weight sharing (v4) is detrimental to the network performance, indicating that SISL learns different weights for pixels further apart. SISL with a dilation rate of 2 shows the best performance. We believe this is because larger dilation rates skip over neighbor information. This skipped information becomes increasingly frequent as the validity mask saturates and consequentially valid pixels are closer to a receptive field with dilation rate of two. We also remove the switch (v5), simply adding the results of the convolutions and output masks, with the results showing the benefit of content-depended dilation.

Removing the mask propagation (v5) and instead propagating validity exclusively at down-sampling layers, similar to Yan \textit{et al.} \cite{Revisiting_Sparsity}, demonstrates the impact of mask propagation on performance. An example of the effect SISL has on the output validity mask is shown in Figure \ref{pic:Binary_Switch}. 

For deeper layers of SISL, when the validity mask is relatively saturated, we experiment with a spatial attention mechanism replacing the binary switch. This mechanism uses a 1$\times$1 convolution followed by the sigmoid activation function, to produce a spatial attention map. This SISL variant improves the networks accuracy, albeit at approximately 660,000 added parameters. Including global context inspired by Qiao \textit{et al.} \cite{DetectoRS} worsens this result. This module summarizes information by channel-wise pooling and learns context through a 1$\times$1 convolution, adding the result to the original features. Most likely, a similar or better performance can be achieved by instead increasing parameters through network depth. SISL v1 will be chosen for our network.

\subsubsection{SI-Residual Bottleneck}
Inspired by further research from He \textit{et al.} \cite{Preactivation_Resnet}, we explore three bottleneck designs. Contrary to the original designs, the SI-convolution normalizes instead of batch normalization. The pre-activation design uses the activation functions before each convolutional layer. Pre-addition moves the last activation function before the added residual. For our ablation study we use the ReLU activation function. The results can be seen in Table \ref{Tb: Ablation-SI-resbn}.

The developed "plain" SI-Residual Bottleneck (v1) significantly outperforms the MA-bottleneck design of Yan \textit{et al.} \cite{Revisiting_Sparsity}, without increasing parameters or memory requirements. We attribute this performance increase to the efficient mask propagation. The "plain" bottleneck also outperforms its pre-activation and pre-addition variants.

\subsection{Benchmark Evaluation}
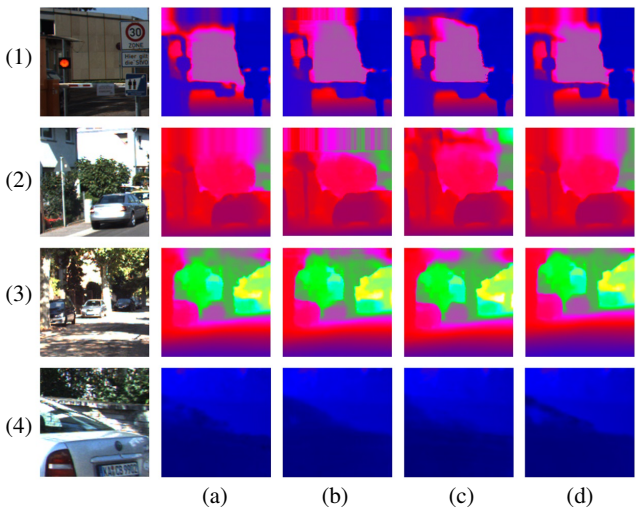
\begin{figure}[t]
	\centering
	\resizebox{\columnwidth}{!}{
		\begin{overpic}[width=\columnwidth, ,tics=5]
			{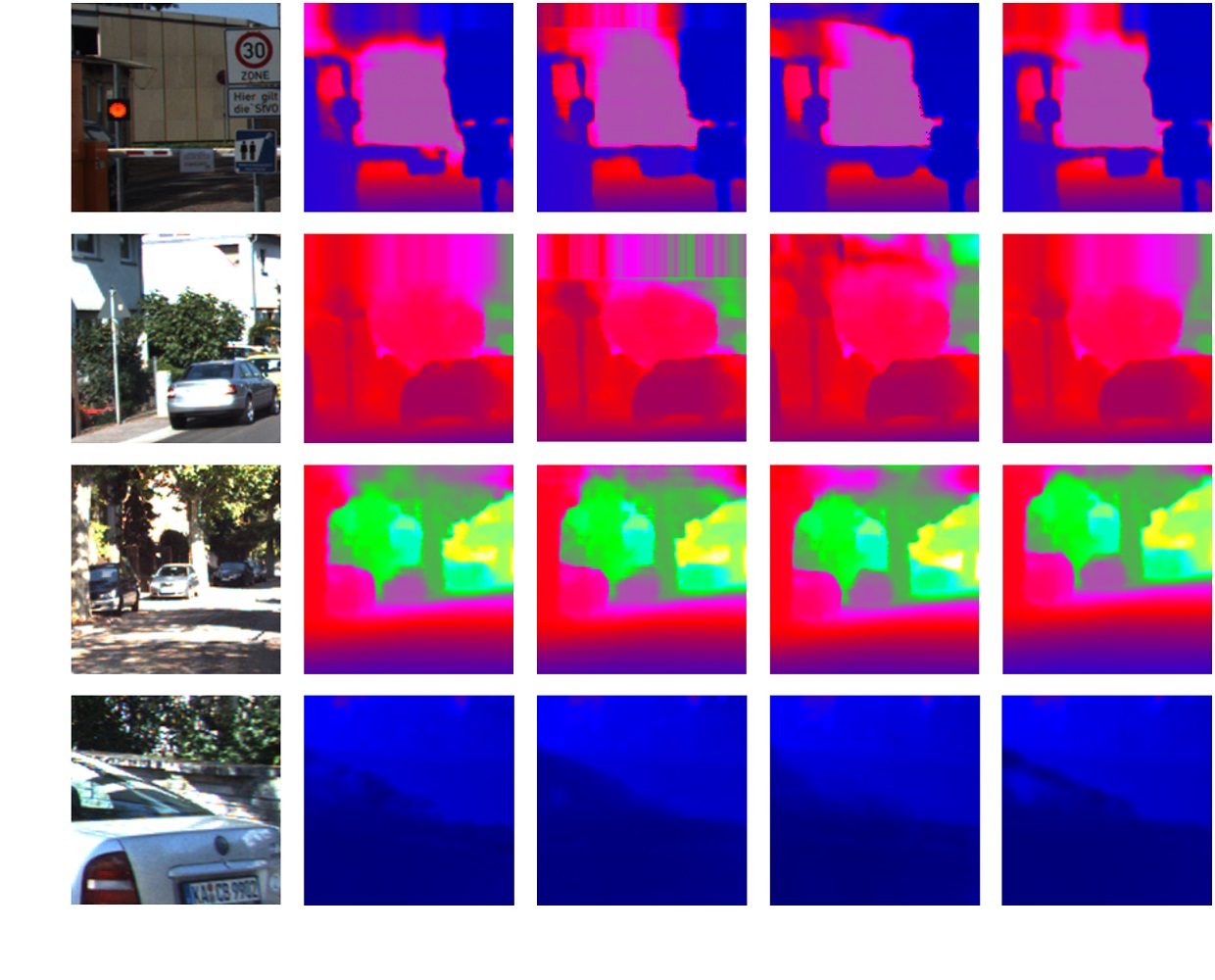}
			\put(2.5,71.5){\makebox[0pt]{\parbox[c]{100pt}{\centering \small (1)}}}
			\put(2.5,52){\makebox[0pt]{\parbox[c]{100pt}{\centering \small (2)}}}
			\put(2.5,34){\makebox[0pt]{\parbox[c]{100pt}{\centering \small (3)}}}
			\put(2.5,13){\makebox[0pt]{\parbox[c]{100pt}{\centering \small (4)}}}
			
			\put(33.5,2){\makebox[0pt]{\parbox[c]{100pt}{\centering \small (a)}}}
			\put(52.5,2){\makebox[0pt]{\parbox[c]{100pt}{\centering \small (b)}}}
			\put(72.5,2){\makebox[0pt]{\parbox[c]{100pt}{\centering \small (c)}}}
			\put(91.25,2){\makebox[0pt]{\parbox[c]{100pt}{\centering \small (d)}}}
			
		\end{overpic}
	}
	\caption{Qualitative comparison of our Dense Validity Mask Network in column (d) to three state of the art solutions: \textit{Revisiting Sparsity Invariant Convolution} \cite{Revisiting_Sparsity} (a), \textit{CrossGuidance} \cite{CrossGuidance} (b), and \textit{PwP} \cite{PWP} (c)}. 
	\label{pic:Kap4:Qualitative}
\end{figure}

\begin{table*}[t]
	\caption{The current KITTI benchmark rankings at the time of writing.}
	\resizebox{\textwidth}{!}{
		\begin{tabular}{@{}lccccccccccc@{}}
			\toprule
			\multicolumn{1}{c}{}                                  &                                                                                 &                                                                                       &                                                                                  &                                       &                                       &                                                                                       &                                                                                         & \multicolumn{4}{c}{\textbf{Benchmark Results}}                                                                                                                                                                                                                                                                               \\ \cmidrule(l){9-12} 
			\multicolumn{1}{c}{\multirow{-2}{*}{\textbf{Method}}} & \multirow{-2}{*}{\textbf{\begin{tabular}[c]{@{}c@{}}RGB-\\ Input\end{tabular}}} & \multirow{-2}{*}{\textbf{\begin{tabular}[c]{@{}c@{}}Encoder-\\ Decoder\end{tabular}}} & \multirow{-2}{*}{\textbf{\begin{tabular}[c]{@{}c@{}}Binary\\ Mask\end{tabular}}} & \multirow{-2}{*}{\textbf{Confidence}} & \multirow{-2}{*}{\textbf{Refinement}} & \multirow{-2}{*}{\textbf{\begin{tabular}[c]{@{}c@{}}Surface \\ Normals\end{tabular}}} & \multirow{-2}{*}{\textbf{\begin{tabular}[c]{@{}c@{}}Additional\\ Dataset\end{tabular}}} & \multicolumn{1}{c|}{\textbf{\begin{tabular}[c]{@{}c@{}}iRMSE\\ (1/km)\end{tabular}}} & \multicolumn{1}{c|}{\textbf{\begin{tabular}[c]{@{}c@{}}iMAE\\ (1/km)\end{tabular}}} & \multicolumn{1}{c|}{\textbf{\begin{tabular}[c]{@{}c@{}}RMSE\\ (mm)\end{tabular}}} & \textbf{\begin{tabular}[c]{@{}c@{}}MAE\\ (mm)\end{tabular}} \\ \midrule
			\multicolumn{1}{l|}{PENet \cite{PE_Net}}                            & \multicolumn{1}{c|}{\checkmark}                                                        & \multicolumn{1}{c|}{\checkmark}                                                              & \multicolumn{1}{c|}{}                                                            & \multicolumn{1}{c|}{\checkmark}              & \multicolumn{1}{c|}{\checkmark}              & \multicolumn{1}{c|}{}                                                                 & \multicolumn{1}{c|}{}                                                                   & {\color[HTML]{5A554E} 2.17}                                                          & {\color[HTML]{5A554E} 0.94}                                                         & {\color[HTML]{5A554E} 730.08}                                                     & {\color[HTML]{5A554E} 210.55}                               \\
			\multicolumn{1}{l|}{FCFR-Net \cite{FCFR_Net}}                         & \multicolumn{1}{c|}{\checkmark}                                                        & \multicolumn{1}{c|}{\checkmark}                                                              & \multicolumn{1}{c|}{}                                                            & \multicolumn{1}{c|}{}                 & \multicolumn{1}{c|}{\checkmark}              & \multicolumn{1}{c|}{}                                                                 & \multicolumn{1}{c|}{}                                                                   & {\color[HTML]{5A554E} 2.20}                                                          & {\color[HTML]{5A554E} 0.98}                                                         & {\color[HTML]{5A554E} 735.81}                                                     & {\color[HTML]{5A554E} 217.15}                               \\
			\multicolumn{1}{l|}{GuideNet \cite{Guided_Conv}}                         & \multicolumn{1}{c|}{\checkmark}                                                        & \multicolumn{1}{c|}{\checkmark}                                                              & \multicolumn{1}{c|}{}                                                            & \multicolumn{1}{c|}{}                 & \multicolumn{1}{c|}{}                 & \multicolumn{1}{c|}{}                                                                 & \multicolumn{1}{c|}{}                                                                   & 2.25                                                                                 & 0.99                                                                                & 736.24                                                                            & 218.83                                                      \\
			\multicolumn{1}{l|}{NLSPN \cite{NL_spatialprop}}                            & \multicolumn{1}{c|}{\checkmark}                                                        & \multicolumn{1}{c|}{\checkmark}                                                              & \multicolumn{1}{c|}{}                                                            & \multicolumn{1}{c|}{\checkmark}              & \multicolumn{1}{c|}{\checkmark}              & \multicolumn{1}{c|}{}                                                                 & \multicolumn{1}{c|}{}                                                                   & 1.99                                                                                 & 0.84                                                                                & 741.68                                                                            & 199.59                                                      \\
			\multicolumn{1}{l|}{CSPN++ \cite{CSPN++}}                           & \multicolumn{1}{c|}{\checkmark}                                                        & \multicolumn{1}{c|}{\checkmark}                                                              & \multicolumn{1}{c|}{}                                                            & \multicolumn{1}{c|}{\checkmark}              & \multicolumn{1}{c|}{\checkmark}              & \multicolumn{1}{c|}{}                                                                 & \multicolumn{1}{c|}{}                                                                   & 2.07                                                                                 & 0.90                                                                                & 743.69                                                                            & 209.28                                                      \\
			\multicolumn{1}{l|}{ACMNet \cite{GraphConv}}                           & \multicolumn{1}{c|}{\checkmark}                                                        & \multicolumn{1}{c|}{\checkmark}                                                              & \multicolumn{1}{c|}{}                                                            & \multicolumn{1}{c|}{\checkmark}              & \multicolumn{1}{c|}{}                 & \multicolumn{1}{c|}{}                                                                 & \multicolumn{1}{c|}{}                                                                   & 2.08                                                                                 & 0.90                                                                                & 744.91                                                                            & 206.09                                                      \\
			\multicolumn{1}{l|}{DeepLidar \cite{DeepLidar_synthdata}}                        & \multicolumn{1}{c|}{\checkmark}                                                        & \multicolumn{1}{c|}{\checkmark}                                                              & \multicolumn{1}{c|}{}                                                            & \multicolumn{1}{c|}{\checkmark}              & \multicolumn{1}{c|}{}                 & \multicolumn{1}{c|}{\checkmark}                                                              & \multicolumn{1}{c|}{\checkmark}                                                                & 2.56                                                                                 & 1.15                                                                                & 758.38                                                                            & 226.50                                                      \\
			\multicolumn{1}{l|}{MSG-CHN \cite{MSG-CHN}}                          & \multicolumn{1}{c|}{\checkmark}                                                        & \multicolumn{1}{c|}{\checkmark}                                                              & \multicolumn{1}{c|}{}                                                            & \multicolumn{1}{c|}{}                 & \multicolumn{1}{c|}{}                 & \multicolumn{1}{c|}{}                                                                 & \multicolumn{1}{c|}{}                                                                   & 2.30                                                                                 & 0.98                                                                                & 762.19                                                                            & 220.41                                                      \\
			\multicolumn{1}{l|}{DSPN \cite{DefSPN}}                             & \multicolumn{1}{c|}{\checkmark}                                                        & \multicolumn{1}{c|}{\checkmark}                                                              & \multicolumn{1}{c|}{}                                                            & \multicolumn{1}{c|}{\checkmark}              & \multicolumn{1}{c|}{\checkmark}              & \multicolumn{1}{c|}{}                                                                 & \multicolumn{1}{c|}{}                                                                   & 2.47                                                                                 & 1.03                                                                                & 766.74                                                                            & 220.36                                                      \\
			\multicolumn{1}{l|}{RGB\&Uncertainty \cite{Sparse_Uncertain}}                 & \multicolumn{1}{c|}{\checkmark}                                                        & \multicolumn{1}{c|}{\checkmark}                                                              & \multicolumn{1}{c|}{}                                                            & \multicolumn{1}{c|}{\checkmark}              & \multicolumn{1}{c|}{}                 & \multicolumn{1}{c|}{}                                                                 & \multicolumn{1}{c|}{\checkmark}                                                                & 2.19                                                                                 & 0.93                                                                                & 772.87                                                                            & 215.02                                                      \\ \midrule
			\multicolumn{1}{l|}{DVMN (ours)}                      & \multicolumn{1}{c|}{\checkmark}                                                        & \multicolumn{1}{c|}{\checkmark}                                                              & \multicolumn{1}{c|}{\checkmark}                                                         & \multicolumn{1}{c|}{}                 & \multicolumn{1}{c|}{}                 & \multicolumn{1}{c|}{}                                                                 & \multicolumn{1}{c|}{}                                                                   & 2.21                                                                                 & 0.94                                                                                & 776.31                                                                            & 220.37                                                      \\ \midrule
			\multicolumn{1}{l|}{PwP \cite{PWP}}                              & \multicolumn{1}{c|}{\checkmark}                                                        & \multicolumn{1}{c|}{\checkmark}                                                              & \multicolumn{1}{c|}{}                                                            & \multicolumn{1}{c|}{\checkmark}              & \multicolumn{1}{c|}{\checkmark}                 & \multicolumn{1}{c|}{\checkmark}                                                              & \multicolumn{1}{c|}{}                                                                   & 2.42                                                                                 & 1.13                                                                                & 777.05                                                                            & 235.17                                                      \\
			\multicolumn{1}{l|}{Revisiting \cite{Revisiting_Sparsity}}                       & \multicolumn{1}{c|}{\checkmark}                                                        & \multicolumn{1}{c|}{\checkmark}                                                              & \multicolumn{1}{c|}{\checkmark}                                                         & \multicolumn{1}{c|}{}                 & \multicolumn{1}{c|}{}                 & \multicolumn{1}{c|}{}                                                                 & \multicolumn{1}{c|}{}                                                                   & 2.42                                                                                 & 0.99                                                                                & 792.80                                                                            & 225.81                                                      \\
			\multicolumn{1}{l|}{CrossGuidance \cite{CrossGuidance}}                    & \multicolumn{1}{c|}{\checkmark}                                                        & \multicolumn{1}{c|}{\checkmark}                                                              & \multicolumn{1}{c|}{}                                                            & \multicolumn{1}{c|}{\checkmark}              & \multicolumn{1}{c|}{}                 & \multicolumn{1}{c|}{}                                                                 & \multicolumn{1}{c|}{}                                                                   & 2.73                                                                                 & 1.33                                                                                & 807.42                                                                            & 253.98                                                      \\
			\multicolumn{1}{l|}{NConv-CNN-L2 \cite{Confidence_L2}}                     & \multicolumn{1}{c|}{\checkmark}                                                        & \multicolumn{1}{c|}{\checkmark}                                                              & \multicolumn{1}{c|}{}                                                            & \multicolumn{1}{c|}{\checkmark}              & \multicolumn{1}{c|}{}                 & \multicolumn{1}{c|}{}                                                                 & \multicolumn{1}{c|}{}                                                                   & 2.60                                                                                 & 1.03                                                                                & 829.98                                                                            & 233.26                                                      \\
			\multicolumn{1}{l|}{SSGP \cite{SSGP_Wasenmueller}}                             & \multicolumn{1}{c|}{\checkmark}                                                        & \multicolumn{1}{c|}{\checkmark}                                                              & \multicolumn{1}{c|}{\checkmark}                                                         & \multicolumn{1}{c|}{}                 & \multicolumn{1}{c|}{\checkmark}              & \multicolumn{1}{c|}{}                                                                 & \multicolumn{1}{c|}{}                                                                   & 2.51                                                                                 & 1.09                                                                                & 838.22                                                                            & 244.70                                                      \\
			\multicolumn{1}{l|}{NConv-CNN-L1 \cite{Confidence_L2}}                     & \multicolumn{1}{c|}{\checkmark}                                                        & \multicolumn{1}{c|}{\checkmark}                                                              & \multicolumn{1}{c|}{}                                                            & \multicolumn{1}{c|}{\checkmark}              & \multicolumn{1}{c|}{}                 & \multicolumn{1}{c|}{}                                                                 & \multicolumn{1}{c|}{}                                                                   & 2.52                                                                                 & 0.92                                                                                & 859.22                                                                            & 207.77                                                      \\
			\multicolumn{1}{l|}{Spade-RGBsD \cite{Sparse_Training_Spade}}                      & \multicolumn{1}{c|}{\checkmark}                                                        & \multicolumn{1}{c|}{\checkmark}                                                              & \multicolumn{1}{c|}{}                                                            & \multicolumn{1}{c|}{}                 & \multicolumn{1}{c|}{}                 & \multicolumn{1}{c|}{}                                                                 & \multicolumn{1}{c|}{}                                                                   & 2.17                                                                                 & 0.95                                                                                & 917.64                                                                            & 234.81                                                      \\
			\multicolumn{1}{l|}{CSPN \cite{CSPN}}                             & \multicolumn{1}{c|}{\checkmark}                                                        & \multicolumn{1}{c|}{\checkmark}                                                              & \multicolumn{1}{c|}{}                                                            & \multicolumn{1}{c|}{}                 & \multicolumn{1}{c|}{\checkmark}              & \multicolumn{1}{c|}{}                                                                 & \multicolumn{1}{c|}{}                                                                   & 2.93                                                                                 & 1.15                                                                                & 1019.64                                                                           & 279.46                                                      \\
			\multicolumn{1}{l|}{Spade-sD \cite{Sparse_Training_Spade}}                         & \multicolumn{1}{c|}{}                                                           & \multicolumn{1}{c|}{\checkmark}                                                              & \multicolumn{1}{c|}{}                                                            & \multicolumn{1}{c|}{}                 & \multicolumn{1}{c|}{}                 & \multicolumn{1}{c|}{}                                                                 & \multicolumn{1}{c|}{}                                                                   & 2.60                                                                                 & 0.98                                                                                & 1035.29                                                                           & 248.32                                                      \\
			\multicolumn{1}{l|}{DFuseNet \cite{DFUSE}}                         & \multicolumn{1}{c|}{\checkmark}                                                        & \multicolumn{1}{c|}{\checkmark}                                                              & \multicolumn{1}{c|}{}                                                            & \multicolumn{1}{c|}{}                 & \multicolumn{1}{c|}{}                 & \multicolumn{1}{c|}{}                                                                 & \multicolumn{1}{c|}{\checkmark}                                                                & 3.62                                                                                 & 1.79                                                                                & 1206.66                                                                           & 429.93                                                      \\
			\multicolumn{1}{l|}{NConv \cite{Confidence_Original}}                            & \multicolumn{1}{c|}{}                                                           & \multicolumn{1}{c|}{\checkmark}                                                              & \multicolumn{1}{c|}{}                                                            & \multicolumn{1}{c|}{\checkmark}              & \multicolumn{1}{c|}{}                 & \multicolumn{1}{c|}{}                                                                 & \multicolumn{1}{c|}{}                                                                   & 4.67                                                                                 & 1.52                                                                                & 1268.22                                                                           & 360.28                                                      \\
			\multicolumn{1}{l|}{ADNN \cite{ADNN}}                             & \multicolumn{1}{c|}{}                                                           & \multicolumn{1}{c|}{}                                                                 & \multicolumn{1}{c|}{\checkmark}                                                         & \multicolumn{1}{c|}{}                 & \multicolumn{1}{c|}{}                 & \multicolumn{1}{c|}{}                                                                 & \multicolumn{1}{c|}{}                                                                   & 59.39                                                                                & 3.19                                                                                & 1325.37                                                                           & 439.48                                                      \\
			\multicolumn{1}{l|}{SI-CNN \cite{uhrig2017sparsity}}                           & \multicolumn{1}{c|}{}                                                           & \multicolumn{1}{c|}{}                                                                 & \multicolumn{1}{c|}{\checkmark}                                                         & \multicolumn{1}{c|}{}                 & \multicolumn{1}{c|}{}                 & \multicolumn{1}{c|}{}                                                                 & \multicolumn{1}{c|}{}                                                                   & 4.94                                                                                 & 1.84                                                                                & 1601.33                                                                           & 481.27                                                      \\ \bottomrule
		\end{tabular}
	}
	\label{Tb:Kap4:Kitti-rankings}
\end{table*}

\subsubsection{Quantitative Evaluation}
DVMN was trained with the same regimen as the ablation studies, but using the AdamW optimizer with a weight decay of 0.01. Its performance was tested on the KITTI benchmark, achieving competitive results (refer to Table \ref{Tb:Kap4:Kitti-rankings}). Our network ranks first among methods using SI-convolution, indicating that our focus on a dense validity mask is beneficial to the information content in such networks.

DVMN has a straightforward architecture, with a single depth map output. Other networks combine multiple depth map predictions into a final map, or refine their result in a second step. DVMN was trained without additional or synthetic data. Among the networks using confidence without refinement or additional data, only \textit{ACMNet} surpasses DVMN, demonstrating that the performance of SI-convolutional networks can be comparable with confidence networks. 

Our network with 2.16M trainable parameters is lightweight in relation to networks using \textit{ResNet-34} \cite{ResBottleneck} or larger backbones \cite{FCFR_Net, CSPN++, PWP, CSPN}. For reference, the original implementation of \textit{ResNet-34} has 21M parameters. Further improvements are conceivable by increasing the network depth or adding a refinement stage.

\subsubsection{Qualitative Evaluation}
To our knowledge, DVMN is the first network with focus on a dense validity mask. We compare the output validity of DVMN to \textit{Revisiting Sparsity Invariant Convolution} \cite{Revisiting_Sparsity}, as a competitive network using SI-convolution, in Figure \ref{pic:Kap4:DVMN_vs_Rev}. Valid output is represented by a while pixel. The additional information gathered by our network is evident. We reach a comparable validity in three layers as \textit{Revisiting} in its third encoder.

Furthermore, we compare our results to three other networks in Figure \ref{pic:Kap4:Qualitative}. In the first example our network (1d) achieves a good reconstruction of the barrier, but excels on the square sign before this barrier. Depth map (1c) produces a rounded sign. In the second example, DVMN has the best completion of the roof-line of the car (2d), with similar observations in (3d) and (4d).

\section{Conclusion}
In this paper we introduced the \textit{Dense Validity Mask Network}, for image guided completion of sparse LiDAR depth maps. We focused on a dense validity mask, increasing the amount of information available in our network. Specifically, we proposed a novel switch layer, using spatially variant and content-depended dilation to gather increased neighbor information and prevent filter degradation. We also integrated sparsity invariant convolution into a residual bottleneck structure, including validity propagation. Our network was evaluated on the KITTI depth completion benchmark and is currently the leading method using a SI-convolution.

\section{Acknowledgements}
This work was funded by the Karl V\"olker Foundation in the project "KI-Fusion". We would also like to thank Dennis Teutscher for his support during the project.


\bibliographystyle{IEEEtran}
\bibliography{IEEEabrv, literatur}

\end{document}